\pdfoutput=1

\documentclass[11pt,dvipsnames]{article}

\usepackage[]{EMNLP2022}

\usepackage{times}
\usepackage{latexsym}

\usepackage[T1]{fontenc}

\usepackage[utf8]{inputenc}

\usepackage{microtype}

\usepackage{inconsolata}

\usepackage{microtype}
\usepackage{graphicx}
\usepackage{tikz}
\usetikzlibrary{patterns,shapes,snakes,calendar,calc,backgrounds,folding,arrows,positioning,math}
\usepackage{calc}
\usepackage{comment}
\usepackage{amsmath,bm} 

\usepackage{multirow}
\usepackage{paralist}
\usepackage{subfigure}
\usepackage{booktabs}
\usepackage[noend]{algpseudocode}
\usepackage{bbding}
\usepackage{stfloats}

\title{Visual Spatial Description: Controlled Spatial-Oriented Image-to-Text Generation}

\author{Yu Zhao$^1$, Jianguo Wei$^1$, Zhichao Lin$^2$, Yueheng Sun$^1$, Meishan Zhang$^3$\Thanks{ Corresponding author}, Min Zhang$^3$ \\
  $^1$College of Intelligence and Computing, Tianjin University \\
  $^2$School of New Media and Communication, Tianjin University \\
  $^3$Institute of Computing and Intelligence, Harbin Institute of Technology (Shenzhen) \\
  \texttt{\{zhaoyucs,jianguo,chaosmyth,yhs\}@tju.edu.cn}, \\ \texttt{\{zhangmeishan,zhangmin2021\}@hit.edu.cn} \\}

\begin{document}
\maketitle
\begin{abstract}
Image-to-text tasks, such as open-ended image captioning and controllable image description, have received extensive attention for decades.
Here, we further advance this line of work by presenting Visual Spatial Description (VSD), a new perspective for image-to-text toward spatial semantics.
Given an image and two objects inside it, VSD aims to produce one description focusing on the spatial perspective between the two objects.
Accordingly, we manually annotate a dataset to facilitate the investigation of the newly-introduced task and build several benchmark encoder-decoder models by using VL-BART and VL-T5 as backbones.
In addition, we investigate pipeline and joint end-to-end architectures for incorporating visual spatial relationship classification (VSRC) information into our model.
Finally, we conduct experiments on our benchmark dataset to evaluate all our models.
Results show that our models are impressive, providing accurate and human-like spatial-oriented text descriptions.
Meanwhile, VSRC has great potential for VSD, and the joint end-to-end architecture is the better choice for their integration.
We make the dataset and codes public for research purposes.\footnote{\url{https://github.com/zhaoyucs/VSD}}
\end{abstract}

\section{Introduction}
Text generation from images is a widely-adopted means for deep understanding of cross-modal data that has received increasing interest of both computer vision (CV) and natural language processing (NLP) communities \cite{DBLP:journals/spm/HeD17}.
Image-to-text tasks generate natural language texts to assist in understanding the scene meaning of a specific image, which might be
beneficial for a variety of applications such as image retrieval \cite{DBLP:conf/aaai/DiaoZML21,DBLP:journals/access/AhmedAAIMC21},
perception assistance \cite{DBLP:conf/eccv/XuYMLHYH18,DBLP:journals/access/ShashiranganaPM21},
pedestrian detection \cite{DBLP:conf/cvpr/HasanLLA021}, and medical system \cite{DBLP:conf/naacl/MiuraZTLJ21}.

\begin{figure}[t]
  \centering
  \includegraphics[width=1.00\linewidth]{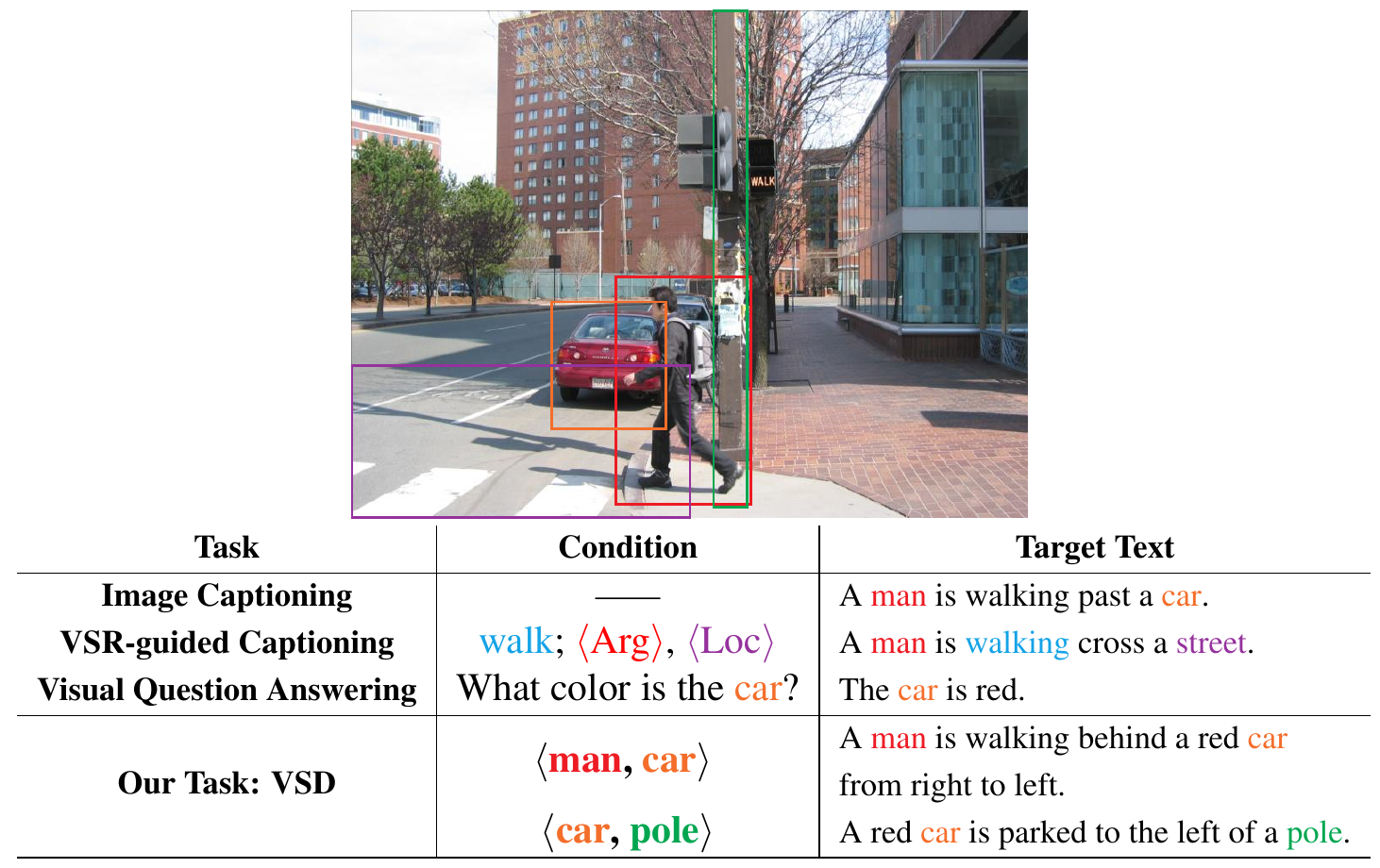}
  \caption{A comparison of three example image-to-text generation tasks and the proposed VSD in this work.}
  \label{fig:0}
\end{figure}

Image-to-text tasks take on various forms when serving different purposes.
Figure \ref{fig:0} illustrates a comparison of three example tasks.
First, the generic open-ended image captioning aims to provide a summarised description that describes an input image and reflects the overall understanding of the image \cite{DBLP:conf/coling/LindhRK20, DBLP:conf/cvpr/VinyalsTBE15,ji2020context}.
Furthermore, the verb-specific semantic roles (VSR) guided captioning \cite{DBLP:conf/cvpr/0016J0021} and visual question answering (VQA) \cite{DBLP:conf/iccv/AntolALMBZP15} are two examples of controllable image description, which produce human-like and stylized descriptions under specified conditions based on a thorough comprehension of the input image \cite{DBLP:conf/cvpr/0016J0021,FeiWRLJ21, DBLP:conf/cvpr/MathewsXH18,DBLP:conf/cvpr/CorniaBC19,DBLP:conf/coling/LindhRK20,DBLP:conf/eccv/Pont-TusetUCSF20, DBLP:conf/eccv/DengDTW20,DBLP:conf/eccv/ZhongWC0L20,DBLP:conf/cvpr/0003COK19,DBLP:conf/cvpr/ChenJWW20,FeiMatchStruICML22,DBLP:conf/emnlp/JhamtaniB18}.
The VSR-guided captioning produces a description focusing on a verb with specified semantic roles,
and the VQA generates a reasoning answer based on a given question.

In this work, we extend the line of controllable image description by presenting the spatial semantics of image-to-text, which is essential but has received little attention previously.
Spatial semantics is a fundamental aspect of both language and image interpretation in relation to human cognition \cite{zlatev2007spatial},
and it has shown great value in spatial-based applications such as automatic navigation, personal assistance, and unmanned manipulation \cite{DBLP:journals/corr/abs-2108-11945,DBLP:journals/corr/abs-2109-15207,DBLP:conf/icra/ZengZSJ18}.
Here, we introduce a new task, Visual Spatial Description (VSD), which generates text pieces to describe the spatial semantics in the image. The task takes an image with two specified objects in it as inputs and outputs one sentence that describes the detailed spatial relation of the objects.
We manually annotate a dataset for inquiry to benchmark this task.

VSD is a typical vision-language generation problem that can be addressed by multi-modal encoder-decoder modeling.
Multi-modal models allow both visual and linguistic inputs and encode them to a joint representation that can learn information from both modal inputs. Moreover, recent studies show that vision-language pretraining can bring remarkable achievements in most image-to-text tasks
\cite{DBLP:conf/nips/LuBPL19,DBLP:conf/iccv/SunMV0S19,DBLP:conf/emnlp/TanB19,DBLP:conf/aaai/ZhouPZHCG20,DBLP:journals/corr/abs-1908-03557,DBLP:journals/corr/abs-2102-10772,DBLP:conf/acl/LiGNXLL0020,xiao2022video}.
Here, we follow these tasks and adopt VL-BART and VL-T5 \cite{DBLP:conf/icml/ChoLTB21} as backbones, which exhibit state-of-the-art performance in vision-language generation.


In particular, a closely-related task, visual spatial relationship classification (VSRC), which outputs the spatial relationship between two objects inside an image, might be beneficial for our proposed VSD. The predefined discrete spatial relations such as ``next to'' and ``behind'', in VSRC should be able to effectively guide the VSD generation.
To this end, we first make a thorough comparison of the connections between VSD and VSRC, which can be regarded as shallow and deep analyses of spatial semantics, respectively, and further investigate the VSRC-enhanced VSD models, performing visual spatial understanding from shallow to deep.
Specifically, we present two straightforward architectures to integrate VSRC into VSD,
one being the pipeline strategy and the other being the end-to-end joint strategy, respectively.


Finally, we conduct experiments on our constructed dataset to evaluate all proposed models.
First, we examine the two start-up models for VSD only with VL-BART and VL-T5.
The results show that the two models are comparable in terms of performance, and both models can provide highly accurate and fluent human-like outputs of spatial understanding.
Second, we verify the effectiveness of VSRC for VSD and find that: (1) VSRC has great potentials for VSD because gold-standard VSRC can lead to striking improvements on VSD;
(2) VSD can be benefited from automatic VSRC, and the end-to-end joint framework is slightly better.
We further perform several analyses to intensively understand VSD and the proposed models.










\section{Related Work}
Image-to-text has been intensively investigated with the support of neural networks in the past years \cite{DBLP:journals/spm/HeD17}. The encoder-decoder architecture is an often considered framework, where the encoder extracts visual features from the image and the decoder generates text for specific tasks.
Early works employ a convolutional neural network (CNN) as the visual encoder and a recurrent neural network (RNN) as the text decoder \cite{DBLP:conf/cvpr/VinyalsTBE15,DBLP:conf/cvpr/RennieMMRG17}.
Recently, the Transformer neural network \cite{DBLP:conf/nips/VaswaniSPUJGKP17},
which is impressively powerful in feature representation learning on both vision and language, has gained increasing interest.
The Transformer-based encoder-decoder models have been adopted in a wide range of image-to-text tasks \cite{DBLP:conf/cvpr/CorniaSBC20,DBLP:conf/nips/HerdadeKBS19,0001RWLJ21}.
These models coupled with visual-language pretraining have achieved the top performance for these tasks \cite{DBLP:conf/nips/LuBPL19,DBLP:conf/iccv/SunMV0S19,DBLP:conf/emnlp/TanB19,DBLP:conf/aaai/ZhouPZHCG20,DBLP:journals/corr/abs-1908-03557,DBLP:journals/corr/abs-2102-10772,DBLP:conf/acl/LiGNXLL0020}.
In this work, we exploit the Transformer-based architecture and two pretrained visual-language models: VL-BART and VL-T5 \cite{DBLP:conf/icml/ChoLTB21}, reaching several strong benchmark models for our task.

Image-to-text can be varied depending on the objective of the visual description.
Image captioning is the most well-studied task, which aims to summarize a given image or to describe a particular region in it \cite{DBLP:conf/cvpr/KarpathyL15,DBLP:conf/cvpr/VinyalsTBE15}.
Several subsequent studies have attempted to produce captions with specified patterns and styles \cite{DBLP:conf/cvpr/CorniaBC19,DBLP:conf/cvpr/0003COK19,DBLP:conf/eccv/DengDTW20,DBLP:conf/eccv/ZhongWC0L20,DBLP:conf/cvpr/ZhengLW19}.
For example, VQA and visual reasoning can be regarded as such attempts, which are conditioned by a specific question directed at the input image \cite{DBLP:conf/iccv/AntolALMBZP15,DBLP:conf/cvpr/AgrawalBPK18,DBLP:conf/cvpr/HudsonM19,DBLP:conf/cvpr/JohnsonHMFZG17}.
The VSR-guided image captioning \cite{DBLP:conf/cvpr/0016J0021} is the most close to our work, which generates a sentence for a particular event in the image with well-specified semantic roles.
Here we focus on spatial semantics instead, generating a description based on the spatial relationship.

Spatial semantics is an important topic in both language and visual analysis.
\citet{DBLP:journals/tslp/KordJamshidiOM11} propose an preliminary study on text-based spatial role labeling.
Later, spatial element extraction and relation extraction from texts are investigated by \cite{DBLP:conf/semeval/NicholsB15}.
\citet{DBLP:conf/semeval/PustejovskyKMLD15} present a fine-grained spatial semantic analysis in texts with rich spatial roles.
Based on the image input, \citet{DBLP:conf/iccv/YangRD19} propose VSRC and benchmark it with a manually-crafted dataset.
The VSRC is actually a shallow task for visual spatial analysis based on a closed relationship set and by using a simple classification schema.
Following, \citet{DBLP:journals/access/ChiouZF21} build a much stronger model on the dataset.
Many studies have exploited spatial semantics to assist other image understanding tasks \cite{DBLP:journals/pr/KimJL21,DBLP:journals/ijcv/WuWHLL21,DBLP:journals/access/CollellDM21,xiao2021boundary,DBLP:journals/ai/PierrardPH21}.
In addition, learning spatial representations from multiple modalities also receives particular attention \cite{DBLP:journals/tacl/CollellM18,DBLP:conf/lrec/DanHR20}.
In this work, we extend image-to-text and propose VSD, which aims for the spatial understanding of the image.

\section{Visual Spatial Description}
\subsection{Task Description}
Formally, we define the task of VSD as follows:
given an image $I$ and an object pair $\left<O_1, O_2\right>$ inside $I$, the VSD aims to output a word sequence $S=\{w_1,...,w_n\}$ to describe the spatial semantics between $O_1$ and $O_2$.
The provided $O_1$ and $O_2$ include both the object tags and their bounding boxes.
In Figure \ref{fig:0}, we would receive ``A man is walking behind a red car from right to left.'' for $\left<man, car\right>$
and ``A red car is parked to the left of a pole.'' for $\left<car, pole\right>$ based on the same input image.
The generated sentences of VSD must encode the spatial semantics between the given two objects, which differs from conventional image-to-text generation.

\subsection{Compared with VSRC}
Noticeably, VSRC is another representative task of visual spatial understanding that decides the spatial relation of two objects in an image.
The relation is chosen from a closed set which is manually predefined.
We can regard VSRC as a shallow analysis task for spatial semantics understanding, while the VSD task can offer a deeper spatial analysis by using the much more flexible output.

In particular, compared with VSRC, VSD has three major advantages. 
First, VSD can offer richer semantics which could be necessary for spatial understanding.
Meanwhile, VSRC only outputs a spatial relation from a closed set in general.
VSD can raise other semantic roles to deepen the spatial understanding beyond the relations, such as predicates and object attributes.
Second, the spatial relations might be overlapped.
For example, the two relationships, ``behind'' and ``to the right of'' might be both correct for VSRC given the ``man'' and ``car'' in Figure \ref{fig:0}.
The newly proposed task VSD can more accurately describe the multiple spatial semantics.
Third, from the viewpoint of downstream tasks, especially the systems that require automatic content-based image indexing or visual dialogue, VSD is more straightforward and adequate to support them.

\subsection{Data Collection}
We build an initial dataset To benchmark the VSD task.
The constructed dataset is extended from a VSRC dataset to facilitate the investigation between VSD and VSRC.
Thus, our final corpus includes both VSRC and VSD annotations.

Our VSRC dataset is sourced from two existing datasets: SpatialSense \cite{DBLP:conf/iccv/YangRD19} and VisualGenome \cite{DBLP:journals/ijcv/KrishnaZGJHKCKL17}.
SpatialSense is a dataset initially constructed for VSRC with nine well-defined spatial relations, namely, ``on'', ``in'', ``next to'', ``under'', ``above'', ``behind'', ``in front of'', ``to the left of'', and ``to the right of''.
The only disadvantage of SpatialSense is its relatively-small scale.
Consequently, we enlarge the corpus with the help of VisualGenome a widely adopted dataset for scene graph generation with annotations in the form of $\langle$subject, predicate, object$\rangle$.
We add the triplets in VisualGenome, whose predicates can be easily aligned with the nine spatial relations in SpatialSense.\footnote{The alignment is achieved by a map, which will be released along with the dataset.}
Accordingly, we can obtain a larger dataset of VSRC.


\begin{figure}[t]
  \centering
  \includegraphics[width=\linewidth]{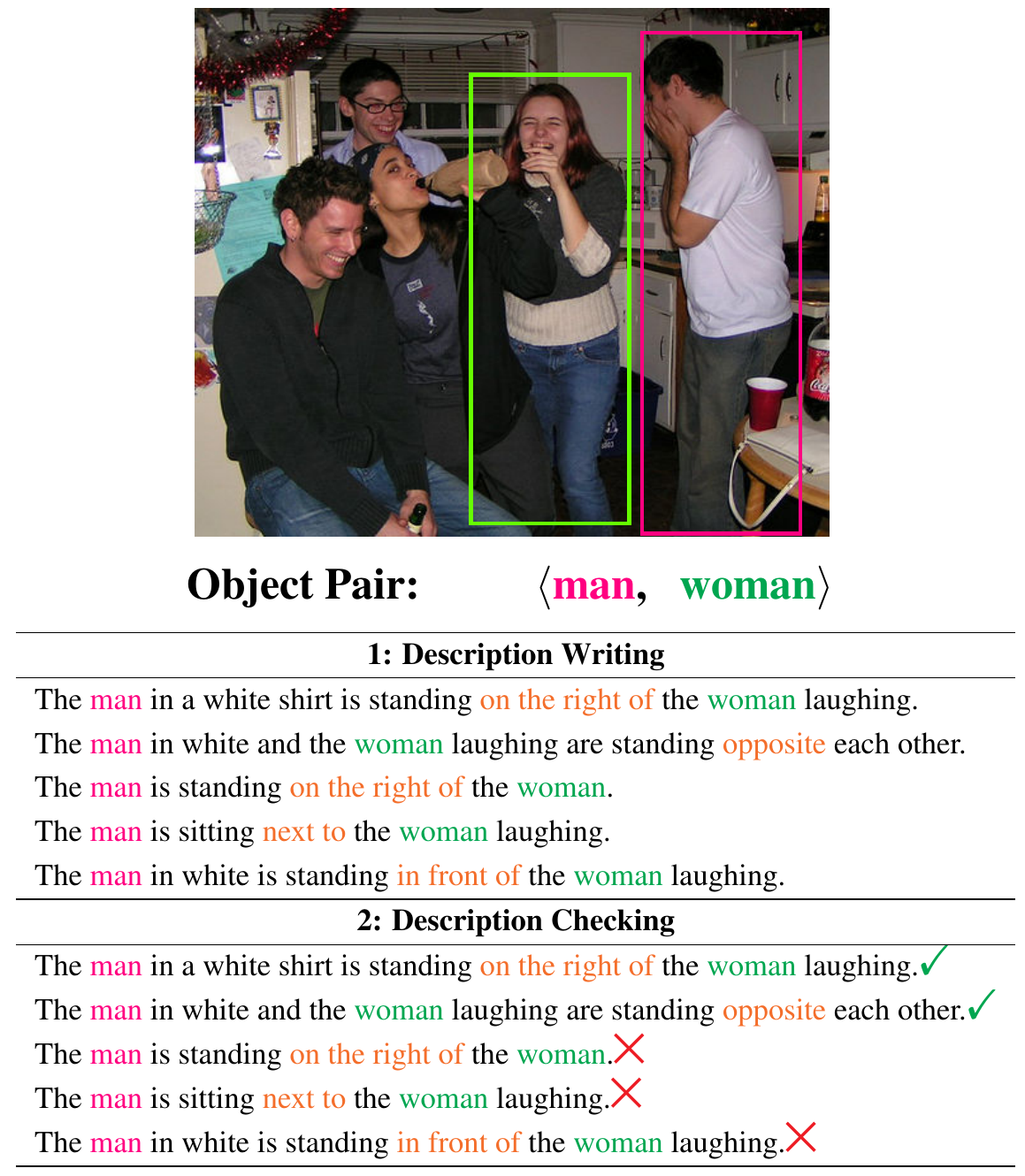}
  \caption{The data annotation flow.}
  \label{fig:annotation}
\end{figure}

We develop a simple visualization tool to facilitate the VSD annotation.
The system randomly assigns the instances to the annotators.
Each instance contains one image and two objects inside it.
The annotators are asked to write text descriptions for the given instance.
We also set up another interface for experts to check the correctness of all annotated sentences and to ensure the quality of these written descriptions.
In the description checking step, the given instances include the image inputs, paired objects, and the written descriptions by the first step. The annotator mainly checks whether the description is valid. The annotation flow is shown in Figure \ref{fig:annotation}.

All annotators we recruited are college students who are native English speakers.
During the preparation, we train the annotators with a guideline and perform two pre-annotation tests from easy to difficult.
In the first test, the annotators are asked to participate in the checking interface, where several well-written descriptions are prepared in advance by experts and various pseudo-ill-conditioned descriptions by intentional word substitutions.
Thereafter, we start the second test to let annotators write the real spatial descriptions.
The annotators are allowed for official annotations only when both tests are passed.
All annotators are properly paid under the open market competition.

Specifically, we have three basic principles mainly in our data annotation guideline as follows:
\begin{compactitem}
    \item The sentence must correctly describe the spatial semantics of the given object pair.
    \item The descriptions can help us correctly locate the exacted objects in the image.
    \item The length of each text description should be limited to no more than 40 tokens.
\end{compactitem}
The annotation submissions with excess invalid annotations (more than 4/100) according the above principles would be returned to the annotators to rework until it reaches the standard. Figure \ref{fig:annotation} shows some examples of invalid annotations. There might be several exceptions, such as, spelling mistakes or mismatches between the image and object tag inputs.
In these cases, annotators should skip and report these instances, leaving them for further discussions by expert.
In the expert-checking step, the remaining invalid and controversial annotations would be be discussed and then finalized.

\setlength{\tabcolsep}{3pt}
\begin{table}[t]
\begin{center}
\begin{tabular}{c|cc|cc}
\toprule
\multirow{3}{*}{Sect.} & \multicolumn{2}{c|}{Input} & \multicolumn{2}{c}{Output}\\ \cline{2-5}
 & \multirow{1}{*}{\#Img}  & {\#OBJ-TAG} & \#SENT &  AvgLEN\\
 \midrule
Train    & 20,490   & 4,506  & 116,791 & 7.35\\
Dev    & 2,927  & 1,416    & 16,823  & 7.33\\
Test & 5,855 & 2,104  & 10,038 & 8.04\\
\bottomrule
\end{tabular}
\caption{The statistics of our constructed VSD dataset.}
\label{tab:data}
\end{center}
\end{table}

Finally, we annotate a total of 29K images with 143K descriptions, wherein 6,591 images are from the SpatialSense with 9,744 descriptions, and the remaining images and descriptions are sourced from VisualGenome.
Furthermore, we randomly split the whole annotated VSD dataset by a ratio of 7:1:2 as training / validation / testing sections.
The statistics of the dataset are shown in Table \ref{tab:data}.


\section{Model}
We exploit the Transformer-based encoder-decoder architecture to accomplish our VSD goal. The architecture can be partially pretrained from ultra-large-scale self-supervised datasets, which makes it capable of obtaining the top performance on a range of image-to-text generation tasks \cite{DBLP:journals/corr/abs-2102-10772,DBLP:conf/acl/LiGNXLL0020,DBLP:conf/emnlp/TanB19,DBLP:conf/eccv/ChenLYK0G0020}.
In this section, we first briefly summarize the adopted model architecture and describe the two well-pretrained models exploited as backbones.

\begin{figure*}[t]
  \centering
  \includegraphics[width=\linewidth]{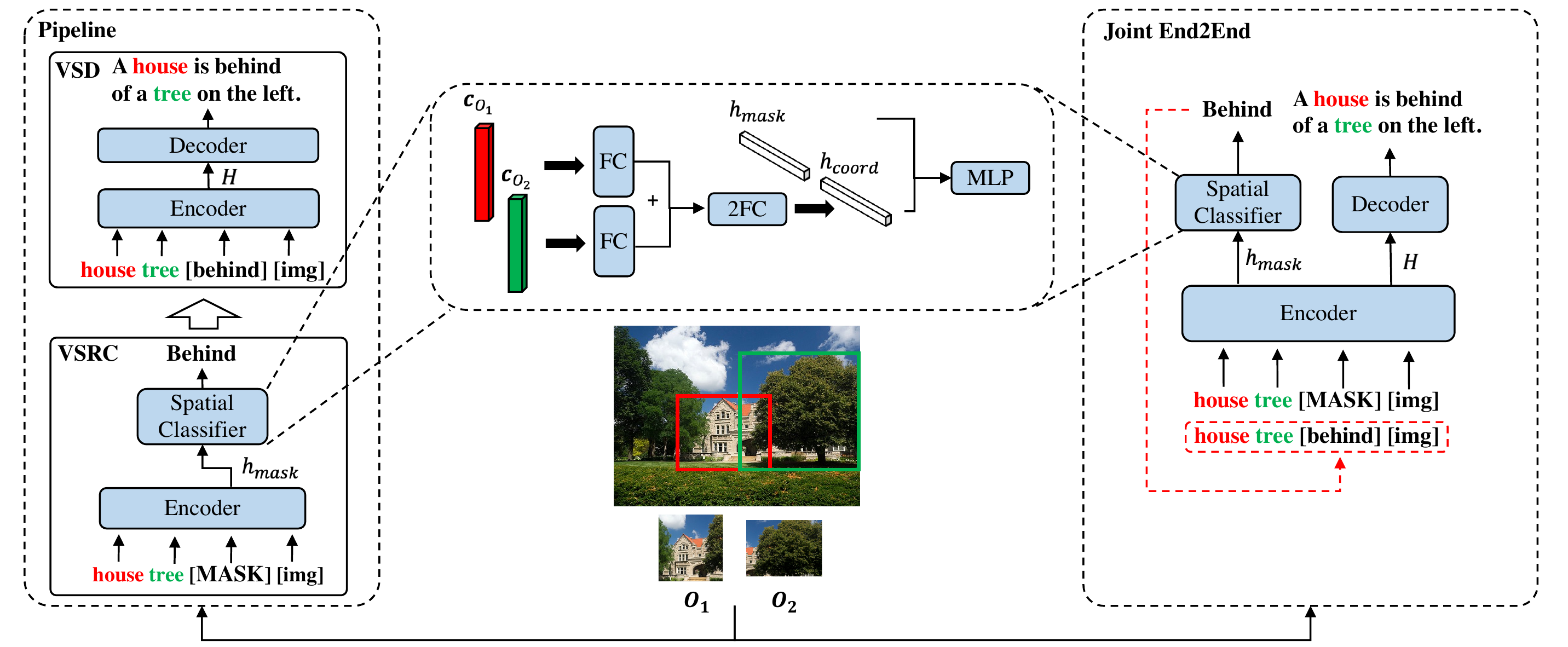}
  \caption{Overview of our pipeline and end-to-end models with VSRC, where FC denotes fully-connected network.}
  \label{fig:2}
\end{figure*}

\subsection{Model Architecture}
The encoder-decoder architecture contains a vision-language (V\&L) encoder and a text decoder.
The encoder takes the combined V\&L inputs to learn a joint feature representation, and the text decoder generates the output sentential words incrementally with the joint representation.

Formally, the VSD input includes: (1) one image $I$ and (2) the inside object pair $\left<O_1, O_2\right>$.
First, we obtain a sequence of visual features by:
\begin{equation}
\begin{split}
  &\bm{F}^{\text{V}} = {\rm VisionExtractor}(I),\\
  &\bm{E}^{\text{V}} = {\rm FC}(\bm{F}^{\text{V}}),\\
\end{split}
\label{eq:visual}
\end{equation}
where $\bm{F}^{\text{V}}$ is obtained by a Faster R-CNN \cite{DBLP:conf/nips/RenHGS15}.
Then a fully-connected (FC) linear transformation layer is used to align the vectorial dimensions between vision and language, leading to $\bm{E}^{\text{V}}$.

Second, a textual embedding layer is used to represent $O_1$ and $O_2$ by their respective textual tags (i.e., $T_{O_1}$ and $T_{O_2}$):
\begin{equation}
\begin{split}
  &\bm{E}^{\text{T}} = {\rm TextEmbed}([T_{O_1}, T_{O_2}]),\\
\end{split}
\label{eq:0}
\end{equation}
where $\bm{E}^{\text{T}} (|\bm{E}^{\text{T}}| = |T_{O_1}| + |T_{O_2}|)$ ($|\cdot|$ indicates the sequence length) is the textual representation of the two objects.

Thereafter that, a Transformer is exploited to produce the final encoder output by the following expression:
\begin{equation}
    \bm{H}={\rm Transformer}([\bm{E}^{\text{T}}, \bm{E}^{\text{V}}]),
\label{eq:1}
\end{equation}
where $\bm{E}^{\text{T}}$ and $\bm{E}^{\text{V}}$ are concatenated and then fed into the Transformer, resulting in $\bm{H}$ which is a sequence of the high-level joint V\&L representations.

We generate a sequence of words incrementally for the decoder,
wherein one word is predicted each step based on the previous context:
\begin{equation}
  \bm{o}_j(\bm{y}|\bm{y}_{i<j}) = {\rm FC}\big({\rm Transformer}(\bm{y}_{i<j} , \bm{H})\big),
\end{equation}
where $\bm{y}_i<j$ denotes the previously generated tokens and $\bm{H}$ represents the encoder outputs.
The decoder is also dominated by Transformer. Thereafter, an FC layer is used to score all candidate words.

We exploit the cross-entropy as objective loss to train the model,
following the majority of sentence generation models \cite{DBLP:conf/acl/LewisLGGMLSZ20,DBLP:journals/jmlr/RaffelSRLNMZLL20}.
During the decoding, we can apply the beam search algorithm to obtain better results.

\subsection{VL-BART and VL-T5}
VL-BART is a well-pretrained model that can be directly applied to our VSD model with an initializing-then-fine-tuning mode.
VL-BART is a standard Transformer-based model similar to our VSD model with a bidirectional joint V\&L encoder and an autoregressive text decoder, which is extended from BART \cite{DBLP:conf/acl/LewisLGGMLSZ20} by importing an extra visual embedding module for the joint encoding.
Before pretraining, VL-BART is partially initialized with BART on the shared parameters,
which is trained on the text-only corpus by corrupting documents and optimizing the model by a reconstruction loss.

VL-T5 is similar to VL-BART on model architecture but differs in that it extends from T5 \cite{DBLP:journals/jmlr/RaffelSRLNMZLL20}.
The T5 model uses relative position embeddings on text representation and is trained on a very different text-only corpus with a span-based reconstruction process.

\section{Enhancing with VSRC}
Our VSD task aims to control image-to-text generation by the aspect of spatial semantics.
If we know the explicit spatial relation by VSRC in advance for the given two objects, then the description generation could be more instructional.
In this section, we introduce two architectures of integrating VSRC into the above-mentioned VSD models.

\subsection{Pipeline}
The pipeline architecture includes two stages.
In the first stage, VSRC is executed to extract spatial relations between the two given objects of the VSD input.
In the second stage, our VSD model adds the spatial relation as one additional textual input to enhance the encoder.
We illustrate the architecture in the left portion of Figure \ref{fig:2}.

Our VSRC model takes the same input as VSD, an image and two objects inside it.
Accordingly, our encoder can be highly similar to that of the VSD model: VisionEmbed and TextEmbed, followed by the Transformer as mentioned in Equations \ref{eq:0} and \ref{eq:1}.
Here, we make a slight modification to adapt the VSRC task.
Specifically, a special [MASK] token is added inside the TextEmbed module:
\begin{equation}
\bm{E}^{\text{T}} = {\rm TextEmbed}([T_{O_1}, T_{O_2}, \text{MASK}]),
\end{equation}
where the updated TextEmbed has been illustrated in Figure \ref{fig:2} by the input depiction of the VSRC.
We only use one vector $\bm{h}_{\text{MASK}}$ from the sequential encoder output $\bm{H}$ for relation classification, which is exactly corresponding to the position of the special [MASK] token.

Before the final-step classification, we follow \cite{DBLP:journals/access/ChiouZF21} to add the bounding box coordinates of the two objects for geometric information.
Each bounding box is converted into a 4-dimensional vector, thus we have $\bm{c}_{O_1}$ and $\bm{c}_{O_2}$ for the two objects, respectively.
Then, we use the following fully-connected (FC) networks sequentially to reach a bounding box representation:
\begin{equation} \label{eq:2}
\begin{split}
  &\bm{\tilde{h}}_{\text{coord}} = {\rm FC}(\bm{c}_{O_1}) + {\rm FC}(\bm{c}_{O_2}),\\
  &\bm{h}_{\text{coord}} = {\rm FC}({\rm FC}(\bm{\tilde{h}}_{\text{coord}})),
\end{split}
\end{equation}
where $\bm{h}_{\text{coord}}$ is the desired bounding box representation.
Finally, we concatenate $\bm{h}_{\text{coord}}$ and $\bm{h}_{\text{MASK}}$ to score candidate spatial relations:
\begin{equation}
\bm{o}^{\text{VSRC}} = {\rm MLP}^{\text{VSRC}}([\bm{h}_{\text{MASK}}, \bm{h}_{\text{coord}}]),
\end{equation}
where ${\rm MLP}^{\text{VSRC}}$ is the classifier for VSRC. In this way, the VSRC task is accomplished.
The middle part of Figure \ref{fig:2} shows the detailed network operation of the classification.

Our VSD task receives three types of inputs from the VSRC output, with additional spatial relation as one supplement compared with the original VSD.
Considering the textual property of the spatial relation, we add this information to the textual embedding of the original VSD encoder:
\begin{equation}
\bm{E}^{\text{T}} = {\rm TextEmbed}([T_{O_1}, T_{O_2}, r_{O_1, O_2}]),
\end{equation}
where $r_{O_1, O_2}$ is the textual expression of the spatial relation between the given objects $O_1$ and $O_2$.
This distinction is the only difference between the VSRC-enhanced and the original VSD models, and the other parts remain the same.

\subsection{End to End}\label{sec:5-2}
The end-to-end model for joint VSRC and VSD is not only more elegant in form, allowing their full natural interactions, but also can avoid the error propagation problem where the VSRC errors may result in further degraded VSD performance.

We adopt multi-task learning (MTL) to achieve the end-to-end goal with a single model.
Figure \ref{fig:2} shows the detailed structure by the right part.
The joint encoder is directly borrowed from the individual VSRC model, resulting in the encoder output $\bm{H}$.
Thus, the input of the joint model is the same as the original VSD model and the VSRC model.
Then, we execute the decoders of VSRC and VSD, achieving the goal of joint learning.

During the training, given the VSRC input (also the joint input) and the VSRC and VSD outputs, we optimize the end-to-end model by the joint loss, which is a weighted addition of the VSRC and VSD losses.
During the inference, we have two strategies for our VSD task.
First, we can use the end-to-end joint model to simultaneously obtain the VSRC and VSD results under the MTL architecture (Figure \ref{fig:2} End2End without the red dashed line).
Second, we can execute the end-to-end model by two rounds, where the first round outputs the VSRC result, and the second round uses the VSRC result to substitute the [MASK] part of the joint encoder, and then executes the VSD part only.
The second strategy is similar to the pipeline architecture, but only a single model is involved.

\section{Experiments}
\subsection{Setup}
\noindent{\bf Implementation Details}
We initialize our encoder-decoder backbone with two pretrained models VL-BART and VL-T5, and follow \cite{DBLP:conf/cvpr/00010BT0GZ18}
to obtain visual region features from Faster R-CNN.
We use the two-round strategy as default for the decoding of the end-to-end models with VSRC, .
We present more model details and hyperparameters in Appendix \ref{ap:setting}.



\noindent{\bf Evaluation}
We report five standard evaluation metrics of the text generation for the VSD task, including BLEU-4 \cite{DBLP:conf/acl/PapineniRWZ02}, ROUGE \cite{lin2004rouge}, METEOR \cite{DBLP:conf/acl/BanerjeeL05}, CIDEr \cite{DBLP:conf/cvpr/VedantamZP15} and SPICE \cite{DBLP:conf/eccv/AndersonFJG16}.
In this work, we use BLEU-4 and SPICE as the primary metrics to evaluate our models, where the former can measure the syntactic quality of the generated descriptions, and the latter emphasizes the consistency with the input scene graphs.
Although CIDEr has been widely-adopted for image-to-text generation as the major metric, it might be unsuitable for our VSD task because it can lower the importance of frequently occurring words closely related to spatial relations by the IDF values.
We conduct each experiment by five times and report the average number.

\usetikzlibrary{fit,shapes.misc}
\newcommand\marktopleft[1]{%
    \tikz[overlay,remember picture]
        \node (marker-#1-a) at (1ex,1.5ex) {};%
}
\newcommand\markbottomright[2]{%
    \tikz[overlay,remember picture]
        \node (marker-#1-b) at (#2,0) {};%
    \tikz[overlay,remember picture,thick,inner sep=3pt]
        \node[draw=red,rounded corners,fit=(marker-#1-a.center) (marker-#1-b.center)] {};%
}
\setlength{\tabcolsep}{6pt}
\begin{table*}[t]
\begin{center}
\begin{tabular}{lcccccc}
\hline
\multicolumn{1}{c}{\multirow{2}{*}{}} & \multicolumn{5}{c}{VSD}  & \multicolumn{1}{c}{VSRC}\\
\multicolumn{1}{c}{}                  & \marktopleft{a1} BLEU-4           & METEOR             & ROUGE             & CIDEr            & \marktopleft{a2} SPICE            & \marktopleft{a3} Acc(\%)\\
\hline\hline
VL-BART                   & 52.71         & 41.96          & 77.57         & 471.21         & 67.83         & -\\ \hline
VL-BART+VSRC-pipeline               & 53.49         & 42.14          & 77.79         & 474.34         & 67.97         & 53.32\\
VL-BART+VSRC-end2end                   & \textbf{53.60}         & \textbf{42.45}          & \textbf{78.15}         & \textbf{476.47}         & \textbf{68.18}         & \textbf{54.53}\\
\textcolor{gray}{VL-BART+VSRC-golden}                   & \textcolor{gray}{72.30}         & \textcolor{gray}{50.90}          & \textcolor{gray}{87.44}         & \textcolor{gray}{578.27}         & \textcolor{gray}{76.59}         & \textcolor{gray}{golden}\\ \hline\hline
VL-T5                    & 52.58         & 41.94          & 77.63         & 472.24         & 67.90         & -\\  \hline
VL-T5+VSRC-pipeline                  & 53.71         & 42.56          & 78.33         & 480.32         & 68.72         & 53.50\\
VL-T5+VSRC-end2end                   & \textbf{54.31}         & \textbf{42.63}          & \textbf{78.38}         & \textbf{481.13}         & \textbf{68.74}         & \textbf{56.36}\\
\textcolor{gray}{VL-T5+VSRC-golden}                   & \textcolor{gray}{72.12}         & \textcolor{gray}{50.95}          & \textcolor{gray}{87.54}         & \textcolor{gray}{579.41}         & \textcolor{gray}{77.29}         & \textcolor{gray}{golden}\\ \hline
\hline
OSCAR$^+$                               & 37.17         & 35.06          & 66.47         & 427.21         & 67.41         & -\\
OSCAR$^+$+VSRC-end2end  & \textbf{38.70}\markbottomright{a1}{0.5em}      & \textbf{35.81}          & \textbf{67.89}         & \textbf{438.28}         & \textbf{67.54}\markbottomright{a2}{0.3em}         & \textbf{57.90}\markbottomright{a3}{0.5em}\\
\hline
\end{tabular}
\caption{The main results of our proposed models on the VSD test dataset, where we implement three types of baseline models (i.e., VL-BART, VL-T5 and OSCAR$^+$), 
and the ones equipped with VSRC supporting.}
\label{tab:result}
\end{center}
\end{table*}

\subsection{Main Results}
Table \ref{tab:result} shows the main results on the test dataset.
The model results based on VL-BART and VL-T5 are reported in two different regions.
The first row of each region shows the performance of our original models.
The base VL-BART and VL-T5 models can achieve impressive performance as a whole, and the two models are generally comparable.
The rows with ``+VSRC--*'' stand for the results of our VSD models with the support of spatial relation.
Meanwhile, the ``VL-T5-*'' models demonstrate better performance under this setting.

To show the potential of VSRC for VSD, we first examine the oracle performance with gold-standard spatial relations as input.
The results are highly exciting, as shown by ``+VSRC--golden'' with the gray numbers.
We can obtain very large improvements over all evaluation metrics based on both VL-BART and VL-T5.
The observation indicates that spatial relation is very useful to our VSD task.
However, using gold-standard spatial relations in real scenarios is impractical.
Thus, it is deserved to investigate the benefits of spatial relations outputted from a VSRC model.

Spatial relations from a VSRC model can be incorporated in two ways, as shown by ``+VSRC-pipeline'' and ``+VSRC-end2end'' in Table \ref{tab:result}.
The two types of models show significant performance decreases compared with that of ``+VSRC-golden''.
Nonetheless, these models can still lead to positive gains on the VSD task by comparing their performance with the basic models without spatial relation information.
In addition, our end-to-end joint models (i.e., ``+VSRC-end2end'') outperform their corresponding pipelines.
If the gain on VSRC is larger, then the increase on VSD is also more significant, indicating that the VSRC performance is the key.

Furthermore, we compare our VL-BART and VL-T5 models with another representative image-to-text model, namely OSCAR$^+$ \cite{DBLP:conf/cvpr/ZhangLHY0WCG21}.
The major difference between our models and OSCAR$^+$ is that OSCAR$^+$ exploits VL-BERT as the backbone, which only contains an encoder.
The spatial relation can effectively improve the OSCAR$^+$ model as well.
Notice that VL-BERT excels at understanding tasks because of its discriminative pretraining benefiting based on sole encoder learning, so we can find that Oscar$^+$+VSRC-end2end can achieve the best VSRC accuracy.
Overall, the OSCAR$^+$ models still obtain lower VSD performance than our suggested VL-BART and VL-T5 models,
demonstrating the advantage of the encoder-decoder pretraining on the VSD task.

\begin{figure}[t]
  \centering
  \includegraphics[width=\linewidth]{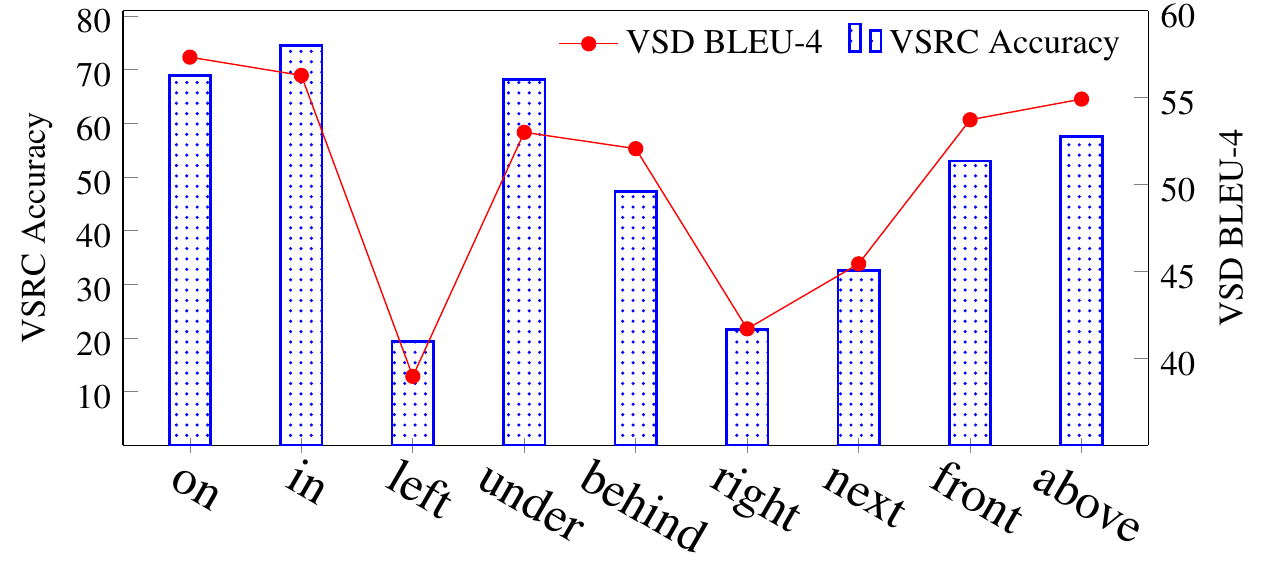}
  \caption{Fine-grained results of the VL-T5+VSRC-end2end model in terms of spatial relations.}
  \label{fig:vrd}
\end{figure}
\subsection{Discussion}

\noindent{\bf Fine-grained Performance by Spatial Relations}
The performance differences among various spatial relations are interesting.
Several particular relations may be more difficult to comprehend within the images.
Figure \ref{fig:vrd} shows the BLEU-4 results across different spatial relations by the VL-T5+VSRC-end2end model,
where the VSRC precisions are also shown for comparison.
Overall, one approximative correction exists between the VSD and VSRC performance by fine-grained evaluation.
Additionally, spatial relations such as ``to the left of'' and ``to the right of'' show significantly lower performance than the others.
The two possible reasons are as follows:
(1) These relations (e.g., including ambiguities by compounds) are visually not as clear as the others, such as ``on'', ``under'', and ``in''.
(2) The distribution of spatial relations is unbalanced.
Although we have paid particular attention to this issue while building our dataset, this problem is still challenging to handle due to the natural characteristic of spatial semantics.
\begin{figure}[t]
 \centering
    \includegraphics[width=0.99\linewidth]{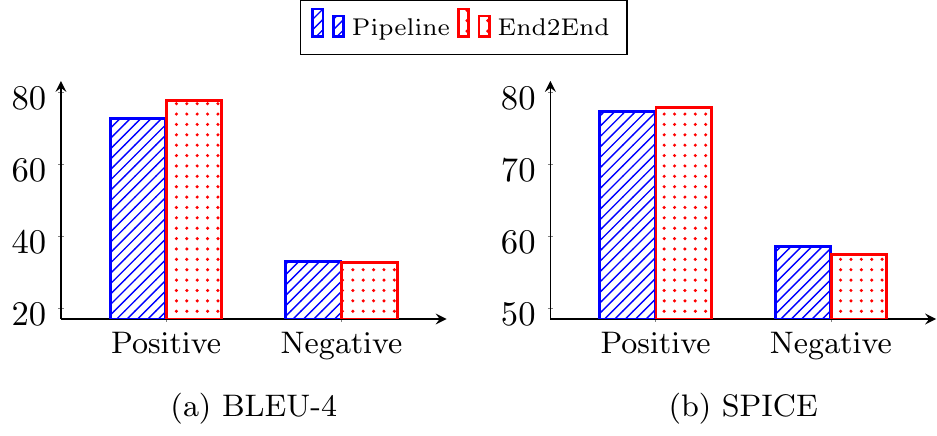}
 \caption{VSD results of VL-T5+VSRC-end2end by Positive and Negative relations predicted from VSRC.}
 \label{fig:pos:neg}
\end{figure}
\begin{figure}[t]
 \centering
    \includegraphics[width=0.99\linewidth]{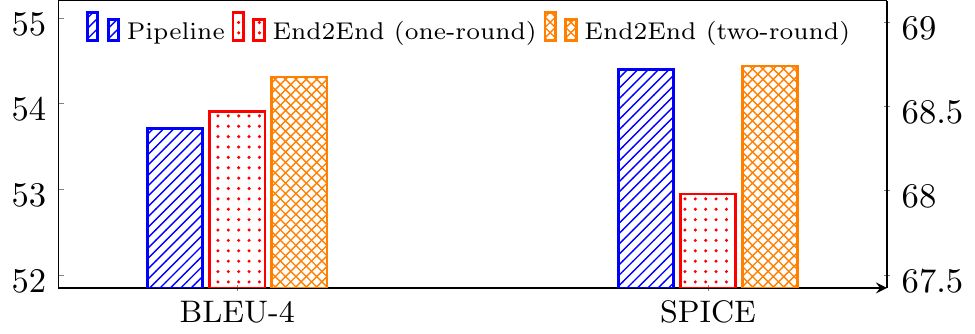}
 \caption{A comparison of the VL-T5+VSRC-end2end model by using one-round and two-round decodings.}
 \label{fig:decoding:step}
\end{figure}

\noindent{\bf Pipeline v.s. End-to-End}
To further understand the disparity between the pipeline and the end-to-end models, we divide the model outputs by the VSRC correctness into two categories and then evaluate the VSD results on them separately.
Concretely, if the VSRC output is correct, then we regard the instance as positive; otherwise, it is negative.
Figure \ref{fig:pos:neg} shows the BLEU-4 and SPICE results by the VL-T5+VSRC-end2end model.
The end-to-end model outperforms the pipeline model on the positive samples while doing the opposite on negative samples.
The obversion is reasonable because the end-to-end model tends to trust its VSRC output by its overall positive influence, thus resulting in downgraded performance when the VSRC outputs are incorrect.
According to the finding,
we can see that the VSRC accuracy is vital  for the final VSD performance.


\setlength{\tabcolsep}{2pt}
\begin{table}[t]
\begin{center}
\begin{tabular}{c|ccc|c}
\toprule
Model & Spatial & Fluency & Location & Avg \\
\midrule
VL-T5(Base)    & 93.2 & 93.8 & 96.1 & 94.4 \\
+VSRC-pipeline & 93.9 & 94.0 & 96.3 & 94.7 \\
+VSRC-end2end & 94.9 & \bf 95.2 & \bf 96.6 &  95.6 \\
+VSRC-golden & \bf 99.3 & 95.0 & \bf 96.6 & \bf 97.0 \\
\bottomrule
\end{tabular}
\caption{Results of human evaluation.}
\label{table:human}
\end{center}
\end{table}

\noindent{\bf Decoding in End-to-end: One Round or Two}
As mentioned in Section \ref{sec:5-2}, we have two strategies for the decoding of the end-to-end models. The two-round strategy is selected by default.
Here, we compare the two decoding strategies based on the VL-T5+VSRC-end2end model.
Figure \ref{fig:decoding:step} shows the results, where the pipeline results are also shown for reference.
The two-round decoding is highly critical for the end2end model, without which the model can even be inferior to the pipeline one.
The possible reason might be that the simple one-round decoding is unable to leverage this advantage even though our MTL architecture for the end2end learning can effectively learn the interactions between the two tasks,
.

\noindent{\bf Human Evaluation}
We perform a human evaluation to better compare the results of our proposed VSD models.
We focus on models with VL-T5 backbone,
and randomly sample 100 test instances of each model for evaluation.
The VSD outputs of each model are scored with the following three measurements:
\begin{compactitem}
\item {\bf Spatial} Correctness: whether the spatial semantics of the generated text is consistent with the image?
\item {\bf Fluency}: whether the generated text is readable and not different from human sentence-making?
\item {\bf Location} Correctness: whether the input objects can be identified from the image according to the generated text?
\end{compactitem}
Each question will be answered by a number from 0 to 1, indicating terrible to perfect. We let three annotators participate in a model-blind evaluation.
Table~\ref{table:human} shows the accumulation scores of over the 100 instances with one decimal place retained. 
Noticeably, the Spatial Correctness is different from the VSRC accuracy in Table \ref{tab:result}, where the former is for human judgement of VSD descriptions and the latter is for a nine-way classification. Generally, the VSRC accuracy can only evaluate one of multiple reasonable spatial relations of the given two objects while the human evaluation is more tolerant and reasonable.
The tendency in performance is consistent with the automatic evaluation, where VSRC can help VSD because it can offer more spatial information, and the end-to-end model is better in utilizing automatic VSRC.
The model with golden VSRC achieves a very high score of 99.3, which is reasonable due to the golden VSRC information of inputs.

\section{Conclusion}
In this work, we introduced a novel image-to-text generation task, namely VSD, aiming to generate text descriptions containing spatial semantics of two objects in an image, and constructed a dataset to benchmark this task.
We adopted the models with Transformer-based encoder-decoder architectures (i.e., VL-BART and VL-T5)  for our task to obtain the baseline results.
Moreover, we proposed to integrate VSRC into our models by pipeline and end-to-end architectures,
enhancing VSD with the support of spatial relations.
The experimental results show that the VSRC-enhanced approach achieves significant progress over our initial models. 
Moreover, the end-to-end models outperform the pipeline ones due to joint learning.

\section*{Limitations}
This work has two major limitations.
The first limitation lies in our dataset.
Our annotated dataset is built on SpatialSence and VG-Relation, aiming to study the relationship between VSRC and VSD.
Under this setting, the variety of the spatial relations is limited to only nine.
In addition, we only annotate one sentence for each instance, which limits the diversity of the description styles.
We plan to continuously improve our dataset with more spatial relations and descriptions as the future work to improve this condition.
The second limitation is that our base models only focus on single spatial relations in this work,
ignoring the compound relations such as ``left'' and ``behind'' concurrently occurring.
To solve this issue, we need to explore more methods to model multiple relations to generate descriptions with richer semantics.
We also leave this aspect to future in-depth studies.

\section*{Ethical Considerations}
We construct a new large-scale image-to-text generation dataset with crowd annotations. All the images of our dataset are sourced from two existing public datasets, SpatialSense and VisualGenome, which are open-access. All the annotators were voluntary participants and can quit at any time. They were informed of the study's goals before giving their express consent. 
All annotators were properly paid by their actual efforts and there is no information related to annotator privacy in the dataset. 

\section*{Acknowledgement}
This work is supported by grants from the National Natural Science Foundation of China (No. 62176180).

\bibliography{anthology,custom}
\bibliographystyle{acl_natbib}

\appendix
\clearpage
\begin{figure*}[tp]
 \centering
    \includegraphics[width=1.00\linewidth]{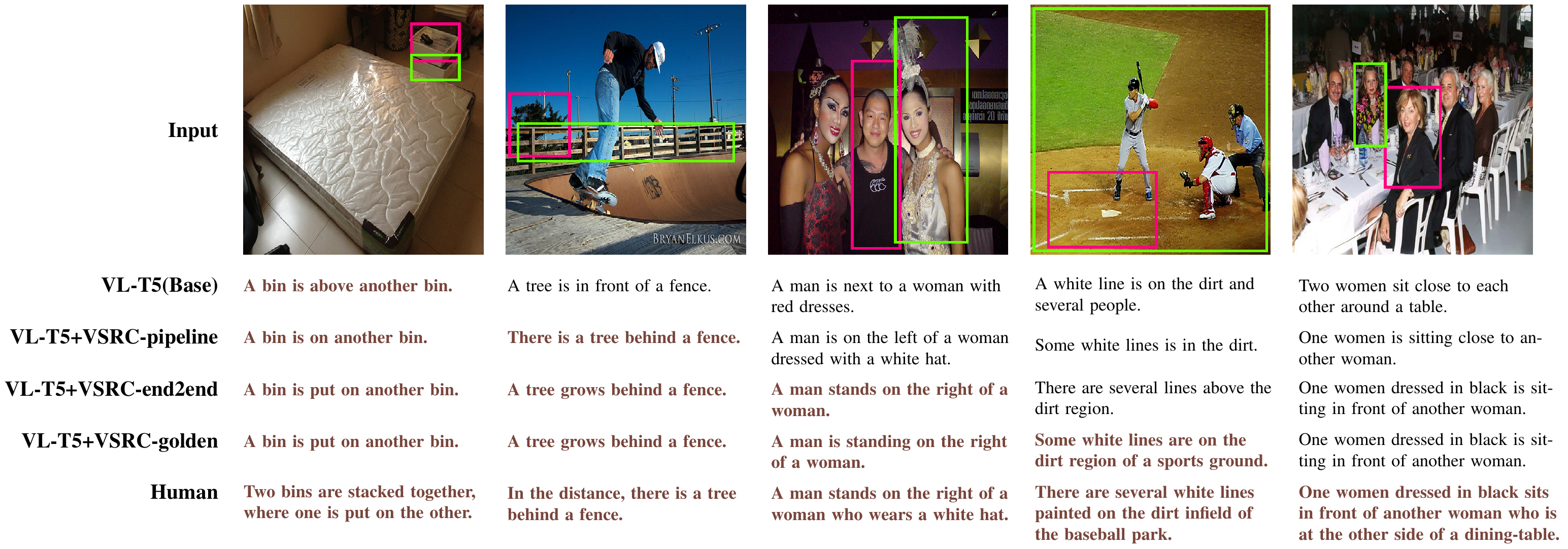}
 \caption{Case studies, where the object in an image marked by the red box is the first object of VSD input, and the bold orange descriptions are regarded as relatively acceptable. }
 \label{fig:B1}
\end{figure*}



\section{Detailed Experiment Settings}
\label{ap:setting}
We adopt the default settings of VL-BART and VL-T5 backbones \cite{DBLP:conf/icml/ChoLTB21}.
In VSRC, the dimension size of the bounding box coordinate features ($\bm{c}_{O_1}$ and $\bm{c}_{O_2}$ in Equation (\ref{eq:2})) is 64 and the dimension of the fully connected layers is set to 1024. For  hyper-parameters, we detail them in Appendix \ref{ap:setting}. We train our models by using the AdamW optimizer \cite{DBLP:journals/corr/abs-1711-05101}, setting the initial learning rate to $5e^{-4}$ and weight decay to $0.01$.
We apply the gradient clipping mechanism by a maximum value of 5.0 to avoid gradient explosion. The training batch size is 16 and the max epoch number is 40.

\section{Case Study}
We show five case studies in Figure \ref{fig:B1} to extensively understand the model outputs. In the first case, all four models (human is the golden answer) are able to output good descriptions because the relation in the image is simple and easy to understand.
In the second case, the VL-T5 (base) model is unable to provide a correct answer, while the other models are all correct due to the benefit from VSRC.
In the third case, the VL-T5+VSRC-end2end and VL-T5+VSRC-golden models output acceptable results, while the other two models fail.
The reason might be that the two models can identify the correct or more important spatial relation between the two objects.
In the fourth case, we can only obtain a correct description by VL-T5+VSRC-golden because the spatial relation is very difficult to recognize by automatic VSRC.
Finally, our VSRC-enhanced VSD model fails in the fifth case even with the golden spatial relation. The reason might be the extreme complexity of this particular input image.



\end{document}